\documentclass[letterpaper, 10 pt, conference]{ieeeconf}  
\IEEEoverridecommandlockouts                              

\usepackage{graphics} 
\usepackage{epsfig} 
\usepackage{mathptmx} 
\usepackage{times} 
\usepackage{cite}
\makeatletter
\let\NAT@parse\undefined
\makeatother
\usepackage{hyperref}
\usepackage{fixmath}
\usepackage{float}
\usepackage{multirow}
\usepackage{amsmath,amssymb, bm}
\usepackage{xcolor}
\usepackage{graphicx}
\usepackage{booktabs}
\usepackage{algorithm}
\usepackage{algorithmic}
\usepackage{amssymb}
\usepackage{mathrsfs}
\usepackage{svg}
\usepackage{makecell}
\usepackage{array}
\usepackage{url}
\usepackage[switch]{lineno}
\newcolumntype{P}[1]{>{\centering\arraybackslash}p{#1}}
\newcolumntype{M}[1]{>{\centering\arraybackslash}m{#1}}
\usepackage{caption}
\usepackage[font=small]{caption}

\newcommand{\thickhline}{%
    \noalign {\ifnum 0=`}\fi \hrule height 1pt
    \futurelet \reserved@a \@xhline
}

\DeclareRobustCommand\onedot{\futurelet\@let@token\@onedot}
\def\onedot{\ifx\@let@token.\else.\null\fi\xspace}
\def\eg{\emph{e.g. }}

\title{\LARGE \bf
NPC: Neural Predictive Control for Fuel-Efficient Autonomous Trucks
}

\author{Jiaping Ren$^{1}$, Jiahao Xiang$^{1,2}$, Hongfei Gao$^{1}$, Jinchuan Zhang$^{1}$, \\ Yiming Ren$^{3}$, Yuexin Ma$^{3}$, Yi Wu$^{4}$, Ruigang Yang$^{1}$, \emph{Fellow, IEEE}, Wei Li$^{1}$ \vspace{-0.5cm} 
\thanks{$^{1}$Inceptio Technology, Shanghai 200082, China, 
        {\tt\footnotesize \{jiaping.ren, hongfei.gao, jinchuan.zhang, ruigang.yang, wei.li\} @inceptio.ai}}%
\thanks{$^{2}$Tongji University, Shanghai 201800, China, 
        {\tt\footnotesize xiang\_jhao@163.com}}%
\thanks{$^{3}$ShanghaiTech University, Shanghai 200120, China, 
        {\tt\footnotesize \{renym1, mayuexin\}@shanghaitech.edu.cn}}%
\thanks{$^{4}$Nanjing University of Posts and Telecommunications, Nanjing 210023, China, 
        {\tt\footnotesize yiw@njupt.edu.cn}}%
}

\begin{document}

\maketitle
\thispagestyle{empty}
\pagestyle{empty}

\begin{abstract}
Fuel efficiency is a crucial aspect of long-distance cargo transportation by oil-powered trucks that economize on costs and decrease carbon emissions. Current predictive control methods depend on an accurate model of vehicle dynamics and engine, including weight, drag coefficient, and the Brake-specific Fuel Consumption (BSFC) map of the engine. We propose a pure data-driven method, Neural Predictive Control (NPC), which does not use any physical model for the vehicle. After training with over 20,000 km of historical data, the novel proposed NVFormer implicitly models the relationship between vehicle dynamics, road slope, fuel consumption, and control commands using the attention mechanism. Based on the online sampled primitives from the past of the current freight trip and anchor-based future data synthesis, the NVFormer can infer optimal control command for reasonable fuel consumption. The physical model-free NPC outperforms the base PCC method with 2.41\% and 3.45\% more significant fuel saving in simulation and open-road highway testing, respectively.

\end{abstract}

\section{Introduction}
\begin{figure*}[t!]
  \centering
    \setlength{\abovecaptionskip}{-0.00cm}
 \includegraphics[width=0.9\textwidth,trim=2 2 2 2, clip]{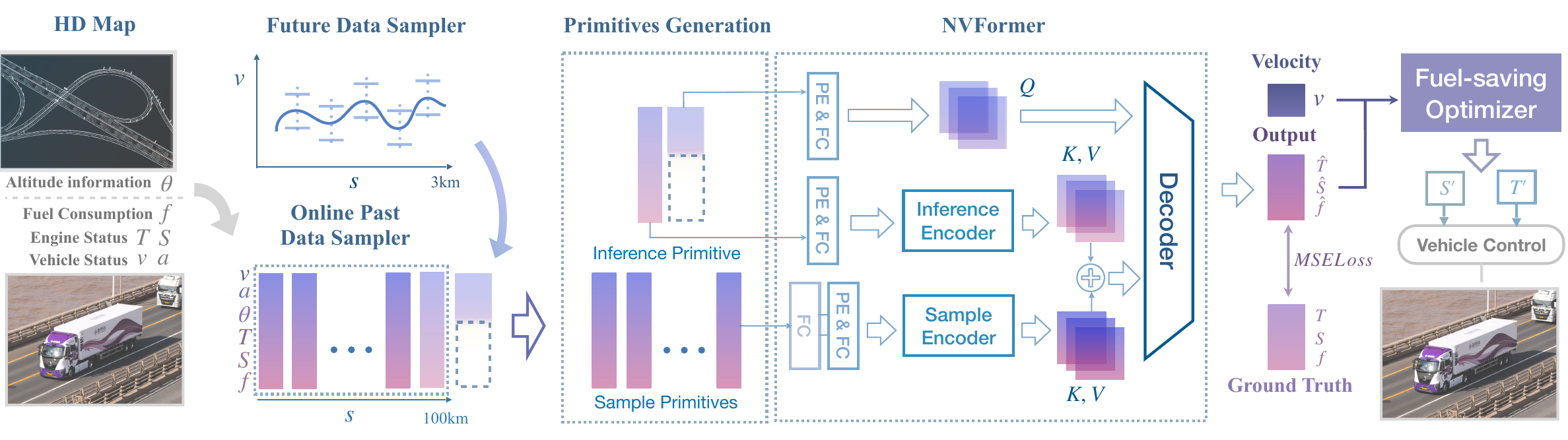}
  \caption{Overview of the Neural Predictive Control (NPC) framework. Both offline and online past data consist of vehicle information from the embedded equipment and altitude information from the HD Map. The proposed NPC samples online past data ($v, a, \theta, T, S, f$) and partial future data ($v, a, \theta$), and generates inference primitive and sample primitives. Inference primitive comprises the latest data sequence along with the partial future data. Sample primitives are distilled from the online past data, excluding the latest data sequence. The NVFormer, which is trained on over 20,000 km offline past data with Mean Squared Error Loss (MSELoss), predicts the missing part ($T, S,f$) of the future data by the inference primitive and sample primitives. Fuel-saving optimizer utilizes the NVFormer to model the relationship between vehicle dynamics, road slope, fuel consumption, and control commands (speed, torque, \emph{etc}.), and output the optimal truck control commands with reasonable fuel consumption.}
  \label{fig:overview}
  \vspace{-0.6cm}
\end{figure*}
Fuel costs are integral to the commercial vehicles' life cycle, particularly in the context of truck operations, where they comprise a substantial fraction of logistics companies' operational expenses. According to ATRI \cite{leslie2022analysis_costs_of_trucking}, fuel costs account for 28\% of truck operational costs. Furthermore, 
road transportation is the major consumer of fossil fuel energy, contributing significantly to worldwide $\rm CO_2$ and pollutant emissions~\cite{eutransportfigures2022}. 
Therefore, improving fuel consumption efficiency can not only reduce freight costs but also benefit the environment. However, the driving skill of human drivers in terms of fuel savings follows a distribution with large diversity and variance~\cite{kamal2009development, van2004driving, walnum2015does_driving_behavior}. Thus, autonomous vehicles, or even advanced driving assistance systems (ADAS) are very promising to improve fuel consumption efficiency with a consistent high-performance driving strategy. 

Recent studies~\cite{zhang2022can, xu2018design_and_comparasion, danielle2018dynamic_eco_driving, schimied2015extension_and_e, yang2016eco} have demonstrated a significant fuel-saving advantage of autonomous vehicles. 
Regarding fuel efficiency for autonomous highway trucking, there is a substantial amount of researches~\cite{organization_of_autonomous_truck_platoon, automotive_platoon_energy_saving, a_review_of_truck_platooning} currently focusing on truck driving strategies. Hu et al.~\cite{minimum_fuel_consumption_strategy_in_ACC} develop an Adaptive Cruise Control (ACC) based on the Pulse-and-Gliding strategy to reduce fuel consumption, with a focus on car-following situations. Dynamic Programming has also been used for minimizing the energy consumption of trucks~\cite{john2020economic_optimal_control} using the local traffic information as the major optimization cost. Ren et al.~\cite{LQR-based_predictive_energy_saving} employ the linear quadratic regulator algorithm for predictive energy saving control based on future road slope, as well as the vehicle dynamic model. Among those fuel-saving solutions, predictive cruise control (PCC)~\cite{lattemann2004predictive, zhang2021fuel, chen2019real_time_pcc, hellstrom2005explicit} is the key line of research possessing the most significant potential for large-scale commercial applications. 

The core of PCC-related works is an optimal control problem (OCP). Solutions for such an OCP using Pontryagin's minimum principle show high performance in fuel saving capability, as well as computation costs~\cite{Saerens2010, SONG2020118064, SHEN2018813, xu2017instantaneous}. However, the performance in the real world is heavily dependent on the dynamic model and the engine/fuel model of the target vehicle. The accuracy of the parameters in the dynamic vehicle model, \eg mass or weight, as well as the coefficient of resistance to drag and rolling, is always overlooked in laboratory experiments and research papers. However, it cannot be ignored in practical applications due to the huge impact on the result of PCC optimization. Technically, these vehicle parameters are estimated using classical system identification solutions\cite{hellstrom2009look, an_enhanced_predictive_cruise_control}. The engine control unit (ECU) of trucks used in our experiments can yield the estimated weight directly. We experimentally find that the ECU's weight estimation error is 8.5\% in our autonomous fleet (747 trips with ground-truth weight from 17.7 tons to 51.85 tons are evaluated). Note that, diverse working conditions in freight tasks, \eg different speed profiles and loads, terrains, weather, and climate would significantly increase the difficulty and error of vehicle parameter estimation. 

To reduce the dependency on the accurate model of vehicle dynamics and engine, we propose a Neural Predictive Control method (see Fig.~\ref{fig:overview}) for fuel-efficient autonomous trucks on hilly roads. NPC is a purely data-driven method, which is free of any physical model for the vehicle. In other words, weight, drag and rolling coefficient, and even the BSFC map of the engine are not needed anymore. Technically, a novel attention-based module NVFormer is designed to implicitly model the relationship between vehicle dynamics, road slope, fuel consumption, and control commands (speed, torque, \emph{etc}.). Based on the online sampled primitives (data from the past of the current freight trip) and anchor-based future data synthesis, the NVFormer can accurately infer control commands for optimal fuel consumption. The physical model free NPC outperforms the base PCC method~\cite{zhang2021fuel} in terms of not only the robustness over varying conditions but also the fuel-saving capability.

The main contributions of this paper are as follows:
\begin{itemize}
\item A pure data-driven NPC is proposed to solve the fuel-optimal control problem. NPC is free of any physical model for the vehicle,~\eg dynamic model and even BSFC map of engine model.

\item A novel attention-based module NVFormer is designed to implicitly model the relationship between vehicle dynamics, road slope, fuel consumption, and control commands (speed, torque, \emph{etc}.). 

\item Compared to the PCC baseline~\cite{zhang2021fuel}, our NPC framework saves 2.41\% fuel in 875.32km close-loop simulation testing and saves 3.45\% in about 145km open-road highway testing with NVFormer which is trained by more than 20,000 km offline real-world vehicle data.
\end{itemize}

\vspace{-0.1cm}

\section{NPC Framework}
In this section, we present the framework of Neural Predictive Control (NPC) for enhancing the fuel efficiency of autonomous trucks. Fig.~\ref{fig:overview} shows the overview of NPC.
The input for NPC includes offline and online data with features vehicle speed ($v$), acceleration ($a$), slope ($\theta$), engine torque ($T$), engine speed ($S$), and fuel consumption ($f$) correspondingly. For one freight trip, complete past data samples and partial future data with features $v$, $a$, $\theta$ are generated by the Online Past Data Sampler and the Future Data Sampler, respectively. Next, NPC yields the inference primitive and the sample primitives. Inference primitive comprises the latest data sequence along with partial future data. Sample primitives are distilled from the online past data, excluding the latest data sequence. We utilize the NVFormer to predict the missing part ($T, S, f$) of the future data by the inference primitive and sample primitives. With NVFormer, the Fuel-saving Optimizer can produce the optimal truck control commands that leverage fuel consumption while meeting transportation constraints.

\subsection{Data Representation}
NPC works on Frenet coordinates due to the predetermined lengths of commercial transportation routes. This coordinate system is extensively employed in autonomous planning tasks \cite{werling2010optimal}. We use $s$ to present distance, and $\{\eta^1,...,\eta^m\}$ to describe the data with $m$ features: $v$,$a$,$\theta$,$T$,$S$,$f$. The data sequence of feature $i$ at a distance $s$ is $\eta^i_s$. The distance at time $t$ is $s_t$. We sample the data every $\Delta s$ m. We use data chunks to represent the sequences of $m$ data. The data chunk at $s_t$ with length $l$ is $\mathbold{\eta}(m, s_t, l) = \{\eta^i_s; i=1, ..., m, s=s_t+\Delta s, ..., s_t+l\Delta s\}$. We denote the $ s \in \{s_t+\Delta s, s_t+\Delta s, ..., s_t+l\Delta s\}$ as endpoint.

\subsection{Online Past Data Sampler}
The Online Past Data Sampler distills important samples from past data with features $v, a, \theta, T, S,f$ in current freight trip. We name such sequential sampled data as \textbf{primitive}, which encompasses representative vehicle states and environment information. The \textbf{inference primitive} comprises the latest data sequence along with the known partial future data ($v, a, \theta$), collectively constituting local features (whether, road friction etc.) in NVFormer. \textbf{Sample primitive} is a data chunk sampled from the online past data, excluding the latest data sequence, which works as global context features (engine properties, vehicle shapes, tie friction characteristics etc.) in NVFormer.
The length of sample primitives is $l_h$ and the latest past data is $\mathbold{\eta}(m, s_t-l_h\Delta s, l_h)$ at time $t$.
When selecting $p$ sample primitives at time $t\in \{t_1, t_2, ..., t_p\}$, the sample primitive $k$ is $\mathbold{\eta}(m, s_{t_k}, l_h)$.

\subsubsection{Composition of Primitive}
The primitive encompasses the following features: $v$, $a$, $\theta$, $T$, $S$, and $f$. Specifically, the $v$, $a$, and $\theta$ represent the value at the endpoint over the distance interval $\Delta s$. The parameters $T$ and $S$ represent the average torque and engine speed, while $f$ denotes the total fuel consumption during this interval.
The NPC model implicitly learns the vehicle's status and dynamic characteristics with the above parameters.
$v$ and $a$ could reflect the vehicle's kinetic information, while $\theta$ encapsulates gravitational potential energy. Additionally, we could generate control commands directly based on $T$ and $S$.

\subsubsection{Sample Primitives Selection}
Sample primitives play a vital role in providing global context features. The vehicle information embedded within sample primitives enhances the prediction capabilities of the model. To provide better global information, sample primitives should encompass various working conditions and vehicle states.
When the distance between the current location and the location of sample primitives exceeds a threshold, the precision of global information drops to unusable. Hence, we establish a maximum effective distance of 100km for sample primitives. To reduce redundancy while maximizing the number of available primitives, we utilize a sliding window with a minimum step size of 1km and a primitive length of 2km. This design ensures the vehicle generates a new set of sample primitives approximately every 100km.

\subsection{Future Data Sampler}
\begin{figure}[h]
\vspace{-0.3cm}
  \centering
    \setlength{\abovecaptionskip}{-0.05cm}
\includegraphics[width=0.85\columnwidth]{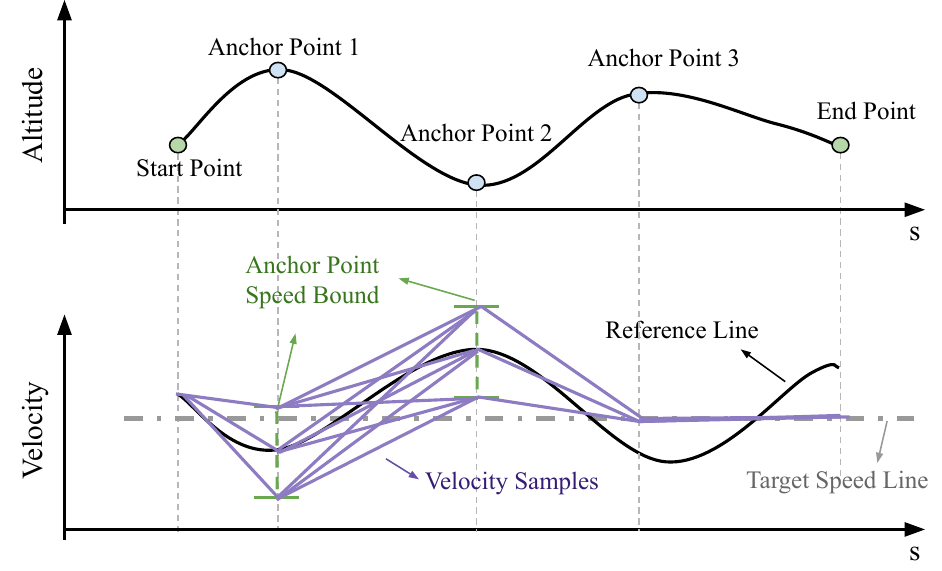}
  \caption{The Future Data Sampler. The s-value of each anchor point refers to the position of the local maximum or minimum of the altitude curvature. The start point, anchor points, and end point together comprise the key points. Velocity curves are sampled within the speed bound of each anchor point.}
  \label{fig:velocity_sample}
  \vspace{-0.3cm}
\end{figure}

The known future data includes the velocity status($v$,$a$) generated by the future data sampler and slope ($\theta$) obtained from the high-fidelity map. We predict $T$, $S$, and $f$ by pre-trained NVFormer and obtain the complete future data. We use the data chunk $\mathbold{\eta}(m, 0, l_f)$ to represent the future data, where $l_f$ is the length of the future data chunk. We sample the speed for the future $l_f\Delta s$ m according to the altitude $z$ (see Fig.\ref{fig:velocity_sample}). Initially, we identify anchor points of the future altitude curve. Subsequently, we calculate speed limits at each anchor point using the reference speed line, which is created based on the current speed and the altitude curve, and the user-defined target speed line. Following this, we generate key points for each sampled speed line and get the future data chunk samples based on the key points.

\subsubsection{Anchor Point Detection} 
An anchor point is a point with local maximum or minimum altitude. We detect anchor points by the condition:
\begin{equation}
\label{eq:anchor_point_condition}
    \frac{\delta z}{\delta s}<\varepsilon,
\end{equation}
where $z$ is the altitude, $\varepsilon$ is the near zero value. For each anchor point, we set the speed bound according to the target speed line defined by the user and the reference speed line created based on the current speed and the altitude curve. The closer, the greater the impact on decision-making at the current location. So, we first sample velocity around anchor points 1 and 2. As shown in Fig. \ref{fig:velocity_sample}, the speed bound of the anchor points 1 and 2 is defined by the reference line. 
Suppose the corresponding reference speed is $v_f$ and the range is $\pm v_d$ m/s, the speed bound is $[v_f^i-v_d, v_f^i+v_d], i=1,2$. The speed of the start point is $v_f^s$, and the speed bounds of the other anchor points (e.g., anchor point 3) and the end point is $v_{t}$, where $v_{t}$ is the target speed. The start point, anchor points, and end point together comprise the key points.
\subsubsection{Speed Sample Generation} 
We evenly divide each speed bound of the first two anchor points (anchor points 1, 2) into $x$ parts, and get $x^2$ sampled speed series. The arbitrary speed series is $\{v_f^s, v_f^{1i}, v_f^{2j}, v_{t}, ...,v_{t}\}$, for $i\in\{1,2,...,x\}$ and $j\in\{1,2,...,x\}$. The speed samples can be generated by interpolating the key points' sampled speed series. Suppose the corresponding $s$ values of the key-points is $\{0, s_{p_1}, ..., s_{p_k},..., l_f\Delta s\}$, $s_{p_k}$ is the $s$ value of the anchor point $k$. For $s_u \in [s_{p_k}, s_{p_{k+1}}]$, the $u$th interpolated speed of the $ij$th speed sample is
\begin{equation}
\label{eq:interpolated_velocity}
    v_u^{ij}=\frac{s_{p_{k+1}}-s_u}{s_{p_{k+1}}-s_{p_k}}v_{p_k}^{ij} + \frac{s_u-s_{p_k}}{s_{p_{k+1}}-s_{p_k}}v_{p_{k+1}}^{ij},
\end{equation}
where $v_{p_k}^{ij}$ is the speed of the $k$th key-point of the $ij$th sampled speed series, $s_u$ is the $s$ value of the $u$th endpoint. Endpoint is the point with $s$ value in $\{\Delta s, ..., u\Delta s,..., l_f\Delta s\}, u \in \{1, 2, ..., l_f\}$.
\subsubsection{Future Data Chunks} 
The future data chunk $\mathbold{\eta}(m, s_t, l_f)$ represents the composition for the future $l_f\Delta s$ m, where $\{\eta^i_s; i=1, ..., m, s=s_t+\Delta s, ...s_t+l_f\Delta s\}$. We divide the total $m=6$ composition into the known composition $v$, $a$, $\theta$ with number $m_f=3$ and the unknown composition $T$, $S$, $f$ with number $m_r=3$. We get velocities $v$ from the velocity sampler, generate acceleration $a$ from $v$, and get slope $\theta$ from the high-fidelity map. With the data chunks, we leverage the pre-trained NVFormer model to predict the unknown composition $T$, $S$, and $f$. Hence, $\eta^i_s=0$ when $i>3$ before prediction.

\subsection{Neural Predictive Model} 
As illustrated in Fig. \ref{fig:overview}, the \textbf{NVFormer} model is built upon an Encoder-Decoder architecture, where the Encoder captures relevant information from sample primitives and inference primitive, and the decoder regresses the predictions. 
NVFormer is a transformer-based model since the multi-head attention mechanism serves a dual purpose. On the one hand, it can attend to the internal relationships within a single primitive. On the other hand, it can establish associations between the global context features of sample primitives and the local features of inference primitive, thereby providing additional information for predictions.

\subsubsection{Encoder Module}
The Encoder comprises a dual ensemble of Transformer Encoders \cite{vaswani2017attention}, namely Sample Former and Inference Former. Sample Former is responsible for processing sample primitives and consists of $n_s$ layer of vanilla Transformer Encoder Layers. Inference Former integrated a stack of $n_i$ Transformer Encoder Layers devoted to processing historical feature series derived from inference primitive. The multi-head attention mechanism in the Transformer Encoder enhances the model's capacity to emphasize pivotal aspects within the input data.

\subsubsection{Decoder Module}
The decoder is structured with $n_i$ layers of Transformer Decoder Layers. It assimilates insights emanating from the Encoder. Concurrently, a sequence mask is employed within the decoder to engage with forthcoming requests, fostering a heightened interdependence between the output and the sequential context of the data.

\subsubsection{Model Workflow}
Through the collaboration of the Future Data Sampler and the Online Past Data Sampler, we obtain a set of $p$ sample primitives and a single inference primitive as inputs, where the future data sequences only encompass the known information ($v$, $a$, $\theta$), while the unknown features ($T$, $S$, $f$) constitute the output of NVFormer.

In a single mini-batch, the shape of sample primitives is $p \times l_h \times m$. These dimensions undergo reduction via a fully connected (FC) layer, resulting in $l_h \times m$. Based on the description above, in an inference primitive, the shape of the latest data is $l_h \times m$. The input future sequence has $m_{f}$ composition, and the sequence's shape is $l_f \times m_f$. For the output sequence, it is $l_f \times m_r$, where $m_r + m_f = m$. 
All data undergo an embedding process before being fed into the attention-based module. This embedding procedure encompasses both positional encoding \cite{vaswani2017attention} and an FC layer. 

During the cross-attention module, the concatenated outputs of the Sample Former and Inference Former are employed as the Key (K) and Value (V) to the decoder module. The future input data sequences, with embedding and a sequence mask, serve as the future request Query (Q) for the decoder. Subsequently, the output of the decoder undergoes the FC layer, leading to the final output.

\subsubsection{Loss Function}

Since NVFormer aims to predict the engine status and fuel consumption, we utilize Mean Squared Error (MSE) as a loss function for training as depicted in Eq. \ref{eq::loss_function}, where $N$ represents the sample quantity, $y_i$  signifies the ground truth values, $\hat{y}_i$ denotes the predicted values.
\begin{equation}
        \mathcal{L}_{MSE} =  \frac{1}{N} \sum_{i=1}^{N} (y_i - \hat{y}_i)^2.
    \label{eq::loss_function}
\end{equation}

\subsection{Fuel-saving Optimizer}

Given the sample primitives, the latest data, and the future data samples, the fuel-saving optimizer selects the best output sequences through the utilization of the pre-trained NVFormer (refer to Algorithm \ref{alg:fuel_saving_planner}). Specifically, the optimizer iterates through each future data sample. For each sample, we employ the pre-trained NVFormer to estimate the curves for torque, engine power, and fuel consumption. Subsequently, we compute the cost by the following equation:

\begin{equation}
\label{eq:planner_cost_function}
    C = w_1\sum_s \eta_s^m + w_2(\left|v_{mean}-v_{target}\right|),
\end{equation}
where $F = \sum_s \eta_s^m, s=\Delta s, ..., l_f\Delta s$ denotes the $m$th feature total fuel consumption. $v_{mean}$ is the average velocity of the future data samples. $v_{target}$ is the target speed. $\left|v_{mean}-v_{t}\right|$ is the speed difference between the $v_{mean}$ and $v_{target}$. The closer between $v_{mean}$ and $v_{target}$, the better to meet the transportation timing requirement. $w_1$ is the weight of fuel and $w_2$ is the weight of velocity. Since we focus more on fuel-saving efficiency, we set $w_1=1.0$ and $w_2=0.1$ in our implementation.
The best future data sample is the one with the lowest cost.
\vspace{-0.4cm}
\begin{algorithm}[ht]
    \caption{Fuel-saving Optimizer}
    \label{alg:fuel_saving_planner}
    \begin{algorithmic}[1]
        \REQUIRE $\mathbold{\eta}(m, s_t-l_h\Delta s, l_h)$,$\{\mathbold{\eta}^i(m, s_t, l_f);i=1,2,...,x^2\}$ \\
        $\bm{T}\gets \bm{0}$, $\bm{S} \gets \bm{0}$,$C_{best}\gets 1000000$
        \FOR{$i = 1$ to $x^2$}
            \STATE Normalize $\mathbold{\eta}(m, s_t-l\Delta s, l_h)$ and $\mathbold{\eta}^i(m, s_t, l_f)$
            \STATE $\{{\eta_{norm}}_s^j; j>3, s=s_t+\Delta s, ..., s_t+l_f\Delta s\} \gets NVFormer(\mathbold{\eta}(m, s_t-l\Delta s, l_f)$,$\mathbold{\eta}^i(m, s_t, l_f))$
            \STATE Denormalize $\{{\eta_{norm}}_s^j\}$ and get $\{\eta_{i,s}^j; j>3, s\in[s_t+\Delta s, s_t+l_f\Delta s]\}$
            \STATE $C_i \gets w_1\sum_{s}\eta^6_{i,s} + w_2(\left|\sum_{s}\eta^1_{i,s}/l_f-v_{target}\right|)$ 
            \IF{$C_i<C_{best}$}
                \STATE $C_{best} \gets C_i$, $\bm{T} \gets \{\eta^4_{i,s}\}$, $\bm{S} \gets \{\eta^5_{i,s}\}$
            \ENDIF \\
        \ENDFOR
        \RETURN $\bm{T}$, $\bm{S}$
    \end{algorithmic}
\end{algorithm}
\vspace{-0.4cm}

\section{Experimental Results}

To validate NPC's fuel-saving efficiency, we train and test the model on an offline dataset derived from real vehicle data. Subsequently, close-loop validation of NPC is performed on both simulation and open-road highway testing.

\subsection{Experimental Results of NVFormer}
We conducted NVFormer training and testing experiments on real-world Inceptio's autonomous truck data. Training the model on offline past data enables it to capture the characteristics of vehicle motion and engine behavior. Additionally, the comparative models included in this part are LSTM and Transformer.

\subsubsection{Dataset}
\begin{figure}[h]
 \vspace{-0.2cm}
  \centering
\setlength{\abovecaptionskip}{-0.00cm}
 \includegraphics[width=\columnwidth]{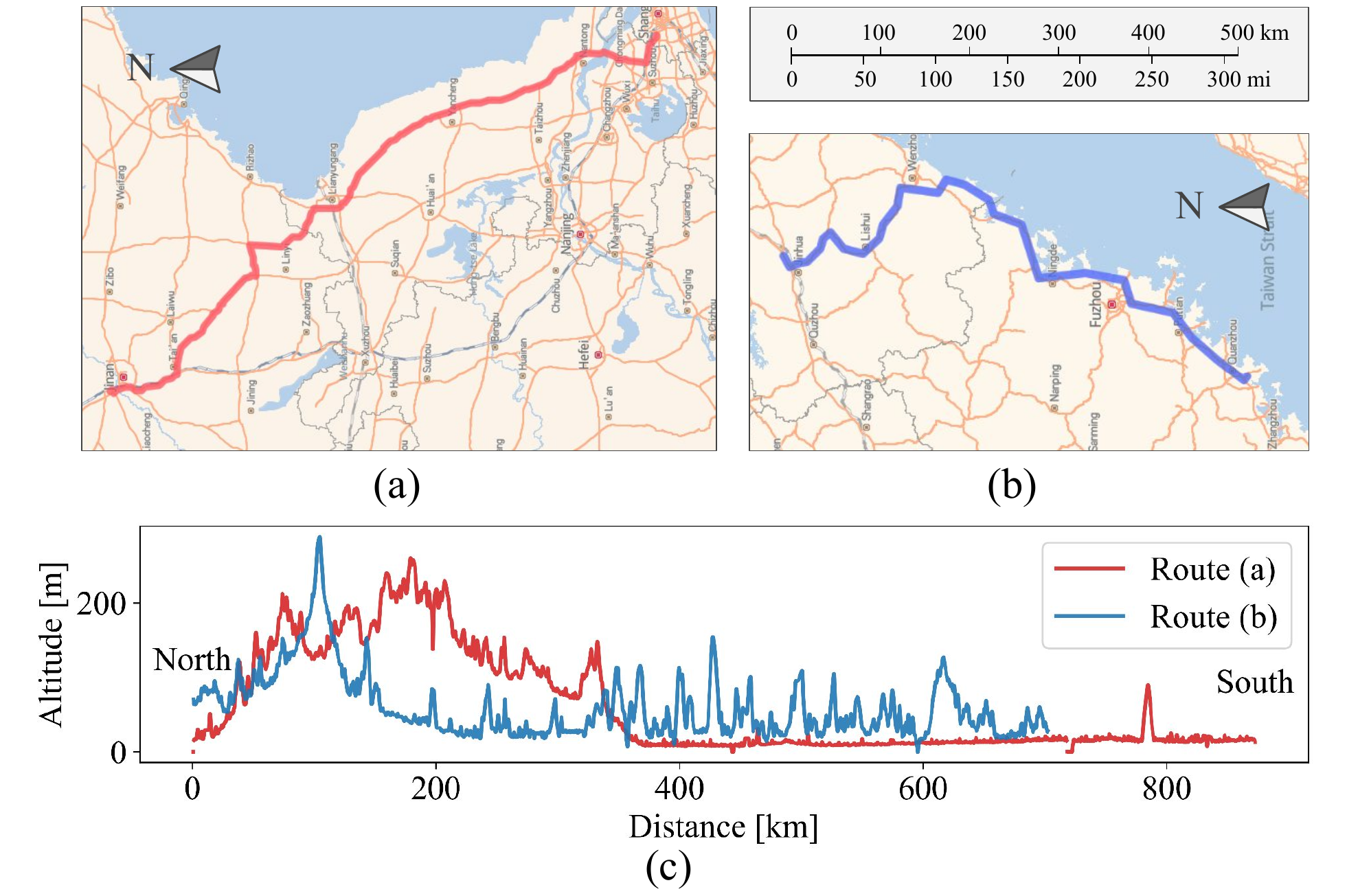}
  \caption{Two of our datasets routes. Route (a) is located in East China, Route (b) is in Southeast China, and (c) represents their altitude variations from north to south.} 
  \label{fig:route_map_with_slope_info}
  \vspace{-0.2cm}
\end{figure}

The existing datasets are not applicable for trucks~\cite{triest2022tartandrive}. Our data is derived from real commercial daily routes of Inceptio's autonomous trucks. All the data was obtained from 4 different vehicles with the same truck series from 26 trips of 4 routes, and the total mileage of all the data is more than 20,000 km. Each vehicle uses GPS, cameras, an inertial measurement unit, and a wire control chassis to record data. The length of each trip is approximately $600\sim850$km, and more than 95\% of the mileage in these trips is on highways. To enhance the model's applicability across various external environments, these routes exhibit varying altitudes, encompassing flat terrains and slopes with different degrees of gradient. For instance, Fig. \ref{fig:route_map_with_slope_info} illustrates two routes from our dataset. As shown in the figure, the northern part of Route (a) consists of mountainous terrain, while the southern part is characterized by plains. In Route (b), the northern section exhibits significant altitude changes, whereas the southern portion has relatively minor elevation variations. Furthermore, our data includes diverse weather conditions and different loads ($18.5\sim36.0$ ton, average 28.2 ton), aiming to increase the diversity of the dataset and enhance the generalization ability of the model. The dataset is used to train the vehicle dynamics, and the drivers' level of driving skill does not matter. The model can acquire inherent vehicle characteristic parameters through the offline past dataset.

By employing a specified step size and point interval $\Delta s = 50 \text{m}$, we employ a sliding window approach to sample multiple sets of data from each individual trip. We generated an offline dataset with about 140,000 primitives in aggregation. All the data is normalized within the range of 0 to 1 before being sent to the model. The ratio of training, validation, and test set is 7:1.5:1.5.

\subsubsection{Baselines and Experimental Settings}
We choose an LSTM-based Encoder-Decoder architecture model (hereinafter referred to as LSTM) and the original Transformer model as baselines for comparative analysis, which could only receive the inference primitive as input. We use 2 encoder layers and 2 decoder layers for all models. Here Transformer and NVFormer are constructed with 8 multi-heads. For NVFormer,  we set the number of sample primitives to 10 with 2 encoder layers in Sample Former. In both LSTM and Transformer architectures, additional FC layers are employed to facilitate the outputs. All models are constructed in the same embedding dimension and compared in data point interval $\Delta s = 50$ m, which leads to input length $l_{h} = 40$ and output length $l_{f} = 60$. 

For all models, we conduct training for hundreds of epochs and select the model with the lowest validation loss after achieving stability for evaluation. A warm-up period of 10 epochs is employed for both Transformer and NVFormer. The initial learning rate for LSTM is $10^{-4}$ and for Transformer and NVFormer is $10^{-5}$. The batch size of training with Adam optimizier \cite{kingma2014adam} is set to 256. All experiments are implemented in PyTorch \cite{paszke2019pytorch} and conducted on two NVIDIA RTX4090 24G GPUs. MAE and MSE are employed as evaluation metrics. 

\subsubsection{Results and Model Analysis}
\begin{table}
    \resizebox{0.9\columnwidth}{!}{
    \begin{tabular}{@{}c|cc|cc|cc@{}}
    \toprule
    \multicolumn{1}{c}{\multirow{2}{*}{Model}} & \multicolumn{2}{c}{\makecell{Torque {[}\%{]}}} & \multicolumn{2}{c}{\makecell{Engine Speed {[}rpm{]}}} & \multicolumn{2}{c}{\makecell{Fuel {[}L/$\Delta s${]}}} \\ \cmidrule(l){2-7} 
    \multicolumn{1}{c}{} & \multicolumn{1}{c}{MAE} & \multicolumn{1}{c}{MSE} & \multicolumn{1}{c}{MAE} & \multicolumn{1}{c}{\makecell{MSE \\ $\times 10^4$}} & \multicolumn{1}{c}{\makecell{MAE \\ $\times 10^{-3}$}} & \multicolumn{1}{c}{\makecell{MSE \\ $\times 10^{-7}$}} \\ \midrule
    LSTM-based & 5.881 & 66.67 & 64.78 & 1.261  & 2.998 & 1.790 \\
    Transformer & 4.792 & 40.44 & 66.17 & 1.170 & 2.974 & 1.780 \\
    NVFormer & \textbf{4.471} & \textbf{34.77} & \textbf{63.05} & \textbf{1.098} & \textbf{2.872} & \textbf{1.622} \\ 
    \bottomrule
    \end{tabular}
    }
\centering
\caption{Prediction performance of LSTM-based model, Transformer and NVFormer. On these three features, NVFormer exhibits the highest level of predictive accuracy.}
\label{tab:baselines_result}
\vspace{-0.6cm}
\end{table}

As demonstrated in Tab. \ref{tab:baselines_result}, in comparison to both LSTM-based and Transformer models, NVFormer exhibits a notable enhancement in predictive accuracy for engine torque. The prediction precision of engine speed and fuel consumption is also significantly elevated through the utilization of NVFormer. Incorporating sample primitives as historical information strengthens the model's capability to predict vehicular states. Within the context of the NPC framework, the provision of more accurate parameter predictions by the model affords NPC a more dependable reference for the vehicle's state.

Our model needs to be deployed onto operational autonomous vehicles. In the real-world scenario, our data points are spaced at $\Delta s = 50$ m intervals, translating to approximately 2 seconds of travel time during high-speed operation. Within the span of these 2 seconds, the NPC Future Data Sampler generates hundreds of velocity sampling lines between two consecutive data points. The NVFormer model, on the other hand, is tasked with processing these velocity lines within the limited time frame. Based on our model architecture design, NVFormer could be separately deployed. Since sample primitives are updated every 100 km, after updating sample primitives, we could use Sample Former to extract features and save the running results locally. After that, only Inference Former and NPC Decoder are running in real-time.
This reduces the required computing power when running the model in real-time.

\subsection{Close-loop Verification of NPC}
We compare our NPC method with PCC \cite{zhang2021fuel} by close-loop verification in simulation and open-road highway. Building upon offline learning, the model within NPC can forecast future vehicle states based on data recorded by the Online Past Data Sampler. All models are executed using the C++ ONNX Runtime API \cite{onnxruntimeapi}.

\subsubsection{Simulation}
\begin{figure}[h]
\vspace{-0.3cm}
  \centering
 \includegraphics[width=0.8\columnwidth]{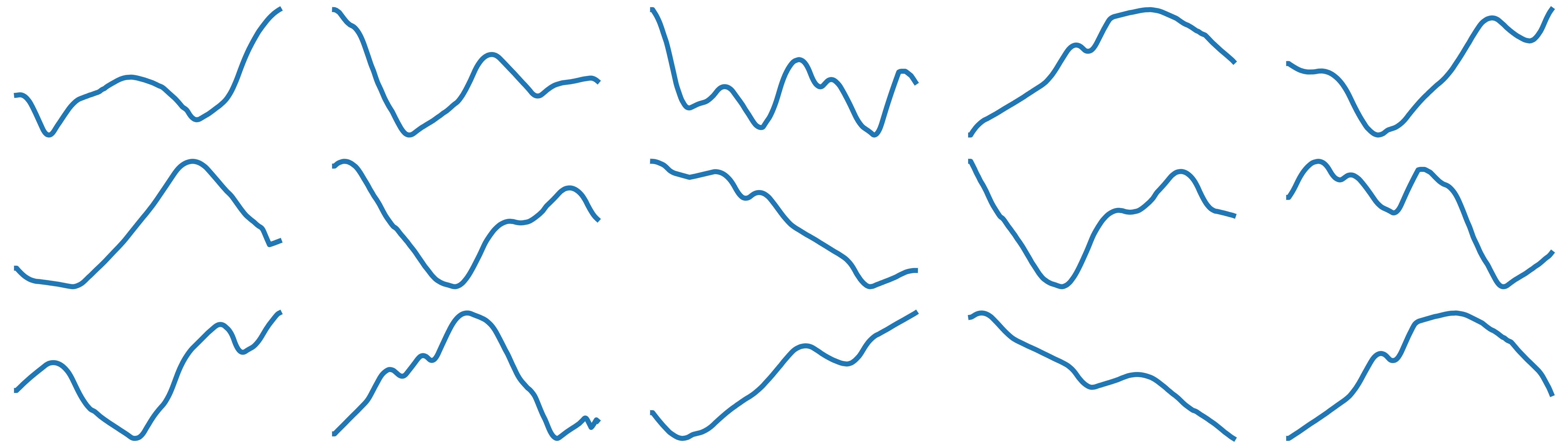}
  \caption{The altitude curves of the 15 different simulation scenarios generated from the real-world map.} 
  \label{fig:slopes}
  \vspace{-0.4cm}
\end{figure}

\begin{figure}[h]
\vspace{-0.4cm}
  \centering
  \setlength{\abovecaptionskip}{-0.15cm}
 \includegraphics[width=0.9\columnwidth]{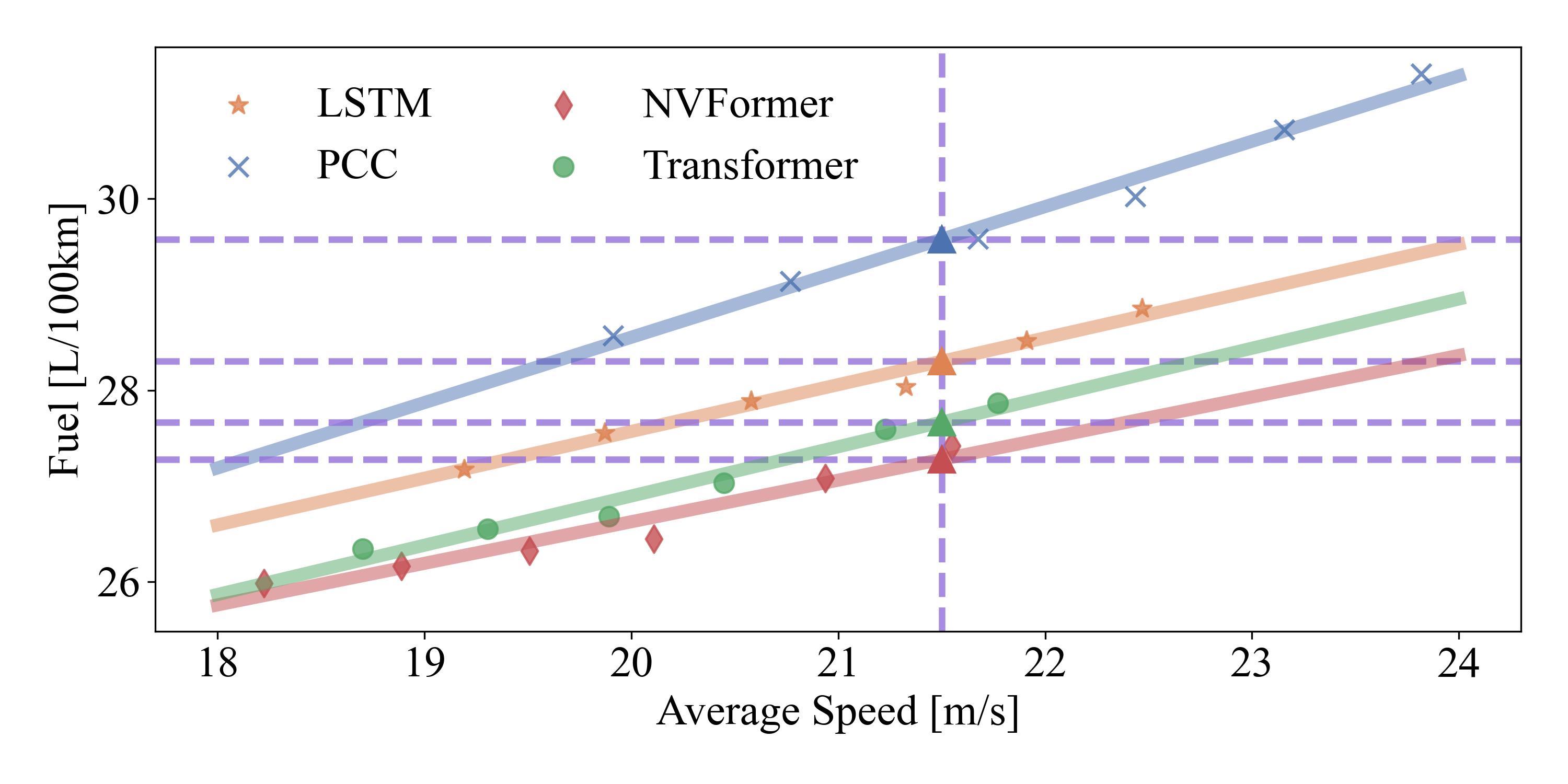}
  \caption{An example of interpolating the fuel consumption at one scenario. We estimate the fuel consumption for each method by interpolating the simulated results at various target speeds from 19.44 m/s to 23.61 m/s.} 
  \label{fig:curve_fitting}
  \vspace{-0.4cm}
\end{figure}

We use simulation to compare PCC, NPC with LSTM model, NPC with Transformer model, and NPC with NVFormer model on real-world road network. All simulation experiments are conducted on the personal computer with 128G memory and Intel Core i9-13900KF processor.\\
\textbf{Scenario and Experimental Settings} We create 15 different simulation scenarios from the real-world system, which include uphill, downhill, flat terrain, undulating road, and so on. Fig. \ref{fig:slopes} shows the altitude of these testing scenarios. Each scenario has a length of nearly 10 kilometers. 
Our simulation employs an in-house well-validated vehicle dynamics model to simulate the motion and performance of a vehicle under various conditions. We simulate fuel consumption interpolated from the BSFC Map~\cite{rajamani2011vehicle}. In each scenario, we test the fuel efficiency of each method with different target speed settings, which include 19.44 m/s, 20.28 m/s, 21.11 m/s, 21.94 m/s, 22.78 m/s, and 23.61 m/s. \\
\textbf{Results} To fairly compare the four methods (PCC, $\text{NPC}_{\text{LSTM}}$, $\text{NPC}_{\text{Transformer}}$ and $\text{NPC}_{\text{NVFormer}}$), we compute the fuel consumption by interpolation from the result. For each method, we conduct experiments with six different target speeds ranging from 19.44 m/s to 23.61 m/s in each scenario. 
We perform interpolation on simulated fuel consumption data at various target speeds to derive a fitting function, which is then used to estimate fuel consumption at 21.5 m/s. The estimated fuel consumption is treated as the fuel consumption of the corresponding method at the corresponding scenario (see Fig.~\ref{fig:curve_fitting}).
We also compute the speed difference, which is the difference between the average speed and the target speed.
As we compute the fuel consumption and the speed difference for all methods at all scenarios, we compute the average fuel consumption $F$, the average speed difference $\Delta v$, and the cost $C_{sim}$ for each method. Similar to Eq.~\ref{eq::loss_function}, the cost function is the weighted sum of the fuel consumption and the speed difference, and
\begin{equation}
\label{eq:cost_function_sim}
    C_{sim} = w_1F + w_2\Delta v,
\end{equation}
where $w_1=1.0$ and $w_2=0.1$. The simulation result is in Tab.~\ref{tab:simulation_result}. Our $\text{NPC}_{\text{NVFormer}}$ has the lowest cost compared to PCC, $\text{NPC}_{\text{LSTM}}$ and $\text{NPC}_{\text{Transformer}}$.

\begin{table}[t]
\setlength{\belowcaptionskip}{-0.5cm}
\centering
\small
\resizebox{0.9\columnwidth}{!}{
\begin{tabular}{c|cccc}
    \toprule
    Method  & \makecell{Fuel \\ {[}L/100km{]}}&  \makecell{Speed Diff. \\ {[}m/s{]}}  & Cost & \makecell{Fuel Saving \\ {[}\%{]}}\\
    \midrule
    PCC & 25.70 & 0.83 & 25.78 & -\\
    $\text{NPC}_{\text{LSTM}}$ & 26.67 & 0.99  & 26.77 & -3.77\\
    $\text{NPC}_{\text{Transformer}}$ & 25.75  & \textbf{0.53} & 25.80 & -0.19\\
    $\text{NPC}_{\text{NVFormer}}$ & \textbf{25.08} & 0.69 & \textbf{25.15} & \textbf{2.41}\\
    \bottomrule
\end{tabular}
}
\caption{Simulation Result. We simulate the four methods (PCC, $\text{NPC}_{\text{LSTM}}$, $\text{NPC}_{\text{Transformer}}$ and $\text{NPC}_{\text{NVFormer}}$) in 15 different scenarios, and set the target speed ranging from 19.44 m/s to 23.61 m/s. Fuel is the average value of all simulations with a total distance of 875.32 km for each method. Speed diff. is the average value of the difference between the average speed and the target speed for all simulations. $\text{NPC}_{\text{NVFormer}}$ has the lowest cost compared to the other three methods and saves 2.41\% fuel compared to PCC.}
\label{tab:simulation_result}
\vspace{-0.2cm}
\end{table}

\subsubsection{Open-Road Highway Testing}
\begin{figure}[h]
\vspace{-0.5cm}
  \setlength{\abovecaptionskip}{-0.1cm}
  \centering
 \includegraphics[width=0.9\columnwidth]{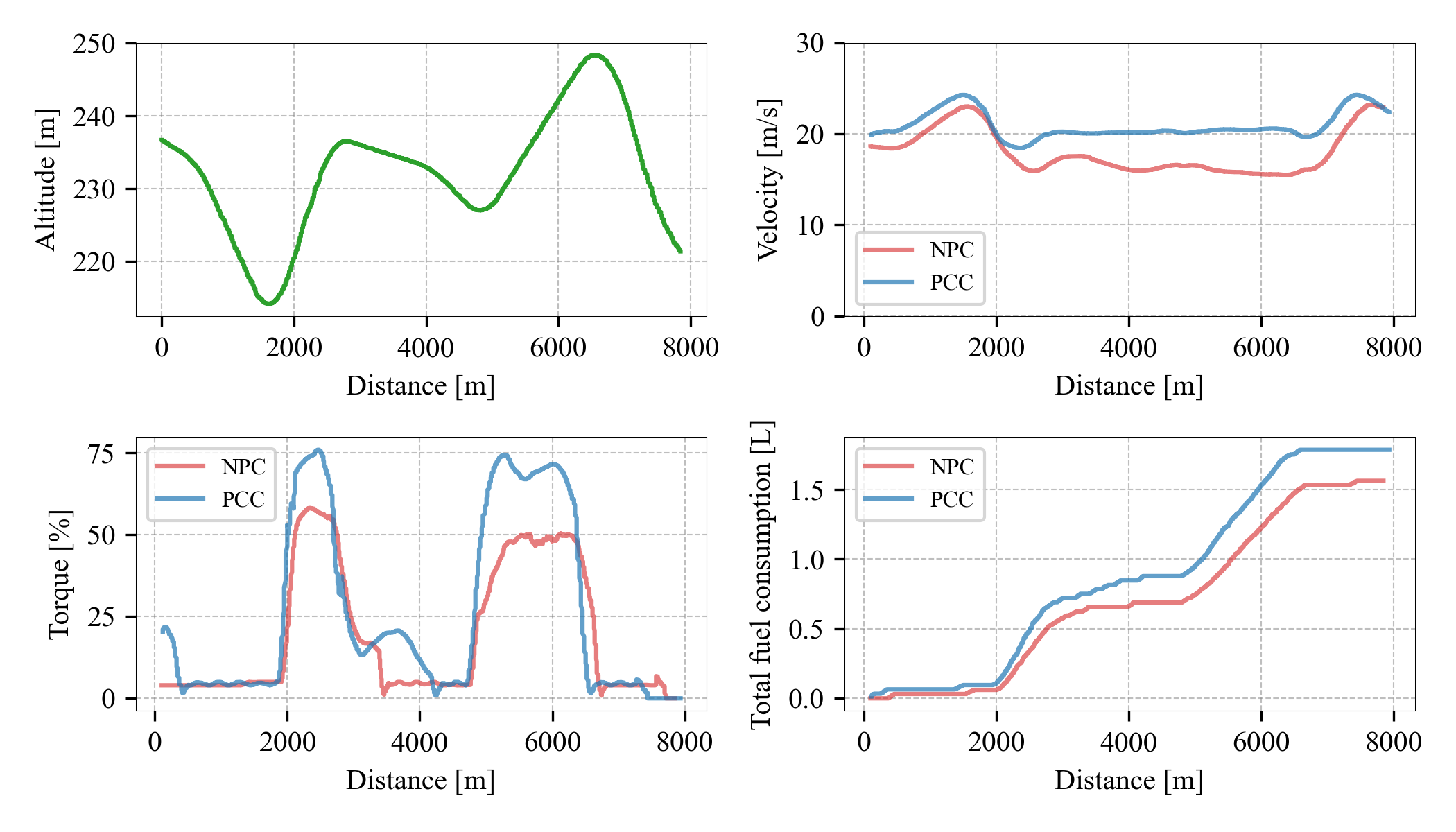}
  \caption{Result of NPC and PCC in part of open-road highway testing. This 8-km long clip covers 2 complete uphill and downhill processes, and one of these three descents is relatively more gradual.} 
  \label{fig:npc_pcc_ort_compare}
  \vspace{-0.5cm}
\end{figure}

The testing was conducted using autonomous driving trucks provided by Inceptio Technology. We made concerted efforts to control test variables, aiming to maintain a consistent testing environment. \\
\textbf{Experimental Settings} 
We conducted testing using the same truck type as an offline dataset for NVFormer. Both methods were subjected multiple times on the same highway section with identical load and driver on the same day. The length of the testing road is approximately 145km, consists of 36\% flat terrain, and 57\% slight uphills and downhills. When testing, the target speed was set to 20.83 m/s. \\ 
\textbf{Results} 
Tab. \ref{tab:real_world_test_result} indicates that NPC with NVFormer demonstrates a fuel efficiency improvement of nearly 1 liter per 100 kilometers compared to PCC in open-road testing. Despite a slightly lower average speed, the cost of NPC is still lower than PCC.
We selected a road segment to conduct an in-depth analysis as shown in Fig. \ref{fig:npc_pcc_ort_compare}. This stretch of road encompasses three descents and two ascents. In the uphill segments, the torque output from NPC is comparatively lower than that of PCC. However, on uphill segments, this velocity differential is regained. In the gentle descent part, NPC maintained a lower torque output, allowing the vehicle to utilize gravitational potential energy to sustain its speed. NPC can achieve global control planning with higher fuel efficiency for the vehicle based on altitude information from a longer distance on the road.

\begin{table}[H]
    \centering
    \begin{tabular}{c|cccc}
        \toprule
        Method  & \makecell{Fuel \\ {[}L/100km{]}} &  \makecell{Avg. Speed \\ {[}m/s{]}} & Cost & \makecell{Fuel Saving \\ {[}\%{]}}\\
        \midrule
        PCC  &  27.50 & 20.07 & 27.58 & -\\
        $\text{NPC}_{\text{NVFormer}}$ & \textbf{26.55} & 19.13 & \textbf{26.72} & \textbf{3.45}\\
        \bottomrule
    \end{tabular}
\caption{Real-world Testing Result. Cost values are calculated by Eq.\ref{eq:cost_function_sim} from simulation testing.}
\label{tab:real_world_test_result}
\vspace{-0.2cm}
\end{table}

\section{Conclusion and Future Work}

In this paper, we present a novel framework to estimate future torque and engine speed series with optimal fuel consumption under transportation timing constraints. To reduce the negative influence of inaccurate vehicle dynamics and engine models, the data-driven NPC method is proposed to take full advantage of a large amount of offline and online vehicle data. NPC performs better than the baseline PCC by the open-loop and close-loop verification. In the future, the efficiency and capability of the network can be enhanced to handle complex traffic on the fly while achieving a safe and fuel-saving autonomous freight trip. We will also learn from large-scale truck operational data to optimize the planner with a deep neural network directly instead of sampling future data to improve the computation efficiency of NPC.

\bibliographystyle{IEEEtran}
\nocite{*}
\bibliography{references}

\end{document}